\documentclass[journal=jctcce,manuscript=article]{achemso}
\SectionNumbersOn
\usepackage[version=3]{mhchem} 
\usepackage[capitalize,noabbrev]{cleveref}
\usepackage{booktabs, multirow}
\usepackage{etoolbox}
\usepackage{graphicx}
\usepackage{amsmath}
\usepackage{amsfonts}
\usepackage{amssymb}

\graphicspath{ {./images/} }

\AtBeginDocument{\singlespacing}

\makeatletter
\patchcmd{\acs@contact@details}{E}{*\,E}{}{}
\makeatother

\author{Liam Harcombe}
\affiliation[ANU]{Australian National University, Canberra, Australia}
\email{liam.harcombe@anu.edu.au}

\author{Timothy T. Duignan}
\affiliation[UQ]{School of Chemical Engineering, The University of \\
Queensland, Brisbane, Australia}
\alsoaffiliation[ORB]{Orbital Materials}
\email{tim@orbitalmaterials.com}

\title{On the Connection Between Diffusion Models and Molecular Dynamics}

\begin{document}








\begin{abstract}
Neural Network Potentials (NNPs) have emerged as a powerful tool for modelling atomic interactions with high accuracy and computational efficiency. Recently, denoising diffusion models have shown promise in NNPs by training networks to remove noise added to stable configurations, eliminating the need for force data during training. In this work, we explore the connection between noise and forces by providing a new, simplified mathematical derivation of their relationship. We also demonstrate how a denoising model can be implemented using a conventional MD software package interfaced with a standard NNP architecture. We demonstrate the approach by training a diffusion-based NNP to simulate a coarse-grained lithium chloride solution and employ data duplication to enhance model performance. 
\end{abstract}


\section{Introduction}

Molecular dynamics (MD) simulations are an indispensable tool for understanding and predicting the behaviour of complex molecular systems, from protein folding and drug discovery to predicting and simulating new materials \citep{durrant2011, larsen2011, merchant2023}. Classical MD models calculate the motion of atoms and molecules by integrating Newton's equations of motion using predefined force fields, enabling extremely fast simulations of large systems. However, these methods trade accuracy for speed, as standard or `classical' force fields approximate interactions using fixed functional forms and cannot fully capture complex phenomena such as bond breaking or electronic effects \citep{brooks2021}. In contrast, ab initio MD calculates the forces at each time step using quantum mechanical methods such as density functional theory \citep{iftimie2005}. Although this approach is highly accurate and can capture more complex phenomena, it is computationally inefficient, making it impractical for simulating large systems or long timescales. A promising approach to overcome these limitations is the use of neural network potentials (NNPs), which are trained on ab initio data to approximate the forces and potential energy surface of a system \citep{batatia2024, batzner2023, behler2007, kocer2022, mailoa2019, merchant2023, park2021, tokita2023, unke2019}. Using machine learning, NNPs can reproduce the accuracy of ab initio methods such as DFT while being orders of magnitude faster, enabling efficient simulations of larger systems and longer time scales. However, a significant downside of NNPs is that they require extensive ab initio calculations to generate the large and diverse datasets needed for training, which can be computationally expensive and time-consuming, particularly for the complex systems we wish to study \citep{artrith2014, smith2017, zhang2018}.

Given the limitations of direct molecular dynamics simulations, recent work has focused on using diffusion models as an alternative to traditional molecular simulation. This approach circumvents the need for force data during training \citep{arts2023, dickstein2015, durumeric2024, pakornchote2024, xie2021, zaidi2022}. These models leverage diffusion processes \citep{ho2020}, where noise is systematically added to molecular systems, and a denoising neural network is trained to learn the reverse process—removing the noise to recover the original system configuration. This approach not only simplifies the data generation pipeline but also reduces computational overhead, paving the way for efficient simulations of complex systems. 
This is most notably demonstrated by Alphafold3 \citep{abramson2024}, a model trained on the Protein Data Bank, which is capable of making remarkably accurate predictions of folded protein structures and other molecules.

A known limitation of Alphafold3, however, is that it can only generate static structures. To overcome this, diffusion models trained on molecular dynamics trajectories are being developed. For example, \citet{jing2024} introduce a generative framework that conditions on initial MD frames to model complete trajectories, capturing not only minimum free energy structures but also the equilibrium distributions, transition paths and dynamics inherent in molecular processes. Their framework leverages the sequential nature of MD data to offer insights into rare event transitions that are inaccessible to equilibrium-only models. Similarly, \citet{lewis2024} demonstrate that generative models trained on extensive MD data can efficiently emulate protein equilibrium ensembles while revealing intermediate conformational states, extending beyond the static predictions of Alphafold3. Their model captures not only the minimum energy structures but also the local fluctuations around the native state.

It is natural to assume that standard NNPs trained on ab initio data and diffusion NNPs trained on noise are fundamentally different, and perhaps even in competition. However, it has been demonstrated in recent years that these two approaches are actually deeply related. Most specifically, it has been shown that learning the score using denoising from an equilibrium distribution of atomic positions is mathematically identical to learning the real physical force field which gives rise to that distribution \citep{zaidi2022}. This has also been demonstrated for real simulations in the case of proteins \citep{arts2023}.

Additionally, this connection is being exploited as a pre-training objective by \citet{zaidi2022} to improve the performance of machine learning models, such as the recently developed Universal neural network potential Orb \citep{neumann2024}.

Here, we aim to elucidate and further demonstrate this connection with a new, simple mathematical derivation using a Taylor expansion. Moreover, we present a practical demonstration that a denoising training scheme can be implemented with a standard molecular dynamics simulation package interfaced with a neural network potential. We show that such a model can generate equilibrium distributions of an electrolyte solution.

\section{Denoising Diffusion Models}
\begin{figure}[ht]
    \centering
    \includegraphics[width=0.8\linewidth]{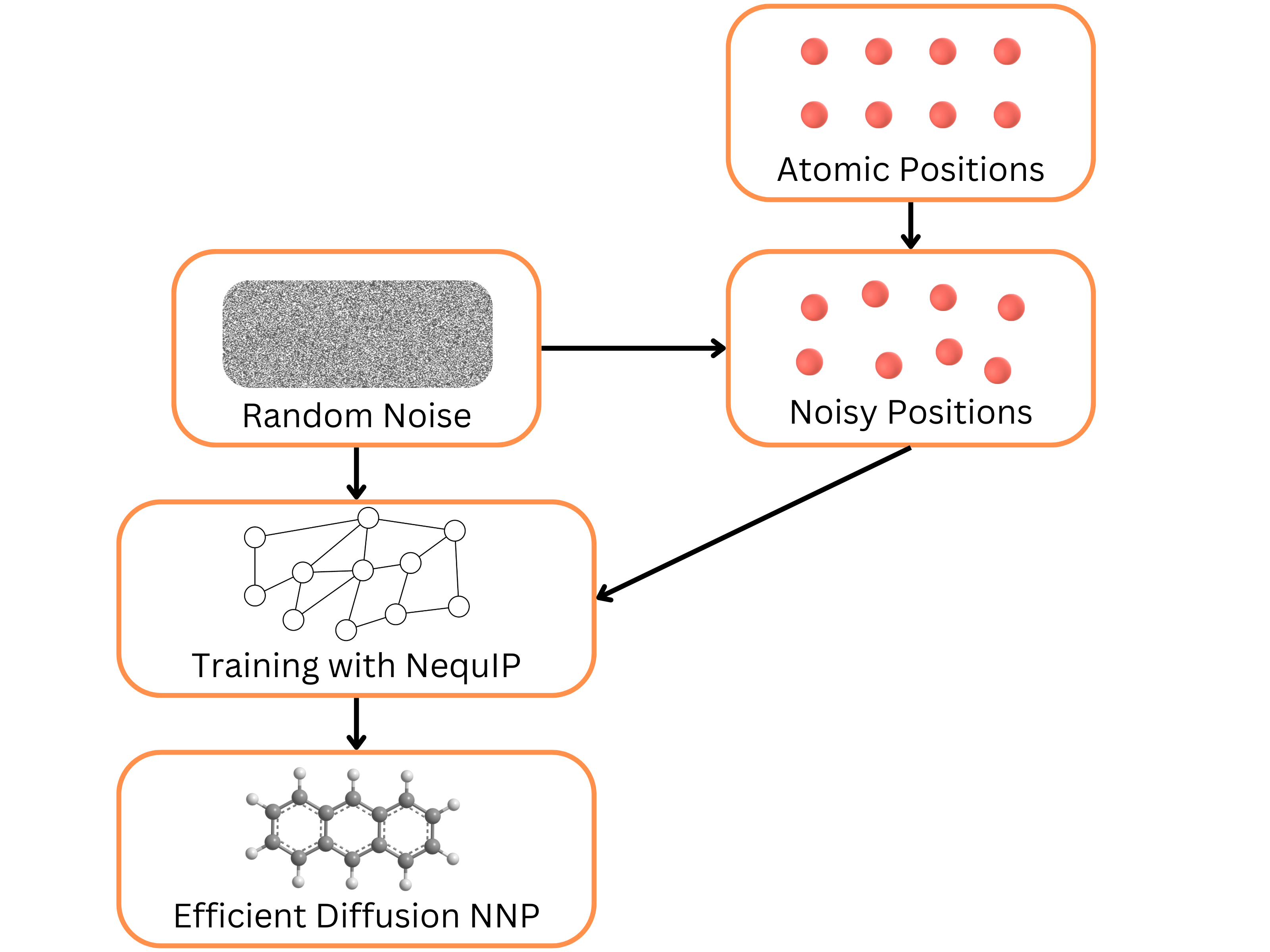}
    \caption{A flow diagram illustrating the process of constructing a NNP using a diffusion-based training scheme.}
    \label{fig:model-diagram}
\end{figure}

The goal of a diffusion model is to generate samples from a probability distribution  $p(\vec{x})$, where $\vec{x}$ is some high-dimensional vector. For example, the vector may represent the pixel values of an image, and $p$ could model the probability that the image depicts a dog, thereby enabling the generation of images of dogs. In our case, we can take the distribution $p$ to be the Boltzmann distribution $\propto e^{-\beta E}$, where $E$ is the energy of the system and $\beta$ is the constant $\frac{1}{k_BT}$. Note that this objective is essentially the same as that of equilibrium MD simulations, and indeed, both methods proceed in very similar ways. 
MD simulations employ algorithms (such as Langevin dynamics or some variant) that are designed to sample from the Boltzmann distribution. In these methods, the forces (i.e., the gradient of the energy) are used to update the particle positions.
Similarly, diffusion models also use Langevin dynamics. In diffusion models, the score $\nabla \log p$ is used in place of the forces. In the case of a Boltzmann distribution, we can see that the $\nabla \log p$ gives rise to the forces, demonstrating the mathematical correspondence between these two approaches. 
The main difference is that in MD we attempt to directly compute the real forces as accurately as possible, whereas for diffusion models the score is learnt from a data set of examples drawn from some distribution. The way this works is by taking vectors $r_i$ of atomic coordinates and vectors $N_i$ of Gaussian noise and training a neural network $F_{\text{DN}}$ on the pairs $(r_i+N_i, -N_i)$. It turns out that this $F_{\text{DN}}$ gives an approximation of the forces on the atoms in the system, which we will derive in Section \ref{subsec:denoise} (previously shown by \citet{zaidi2022}). 
A schematic for this design is shown in Figure \ref{fig:model-diagram}.

The intuition behind why $F_{\text{DN}}$ provides the forces is to imagine the coordinates $r_i$ representing atoms in a stable state. Adding the noise vectors $N_i$ to the $r_i$ pushes the atoms into a more unstable configuration (on average). Thus, learning $-N_i$ is essentially learning how to move the atoms back into the more stable state, which is exactly what forces do.

One caveat is that in general, diffusion denoising models use a time-dependent score. We will ignore this for the purposes of demonstrating the connection to molecular dynamics, but discuss it more below as it is an important feature of these models. 

\subsection{Denoising} \label{subsec:denoise}
To clarify exactly how the denoising process works to generate the correct force field, we provide a simple argument that demonstrates why this is true in one dimension, which can be easily generalised to arbitrary dimension. Let $\mathcal{N}$ be the distribution from which we take our noise vectors $N_i$, usually the normal distribution with mean $r'$ and standard deviation $\sigma$:
\[
\mathcal{N}(r, r', \sigma) = \frac{1}{\sigma \sqrt{2\pi}}e^{-\frac{1}{2}\left(\frac{r-r'}{\sigma }\right)^2}.
\]
With $\rho$ as density, we write the force computed with the denoising algorithm as
\[
F_{\text{DN}} = \langle (r'-r)\rangle _{\rho'(r)\mathcal{N}(r, r', \sigma)} = \frac{\int_{-\infty}^\infty (r'-r)\rho(r')\mathcal{N}(r, r', \sigma)dr'}{\int_{-\infty}^\infty\rho(r')\mathcal{N}(r, r', \sigma)dr'}.
\]
The distribution we average over is the product of the true distribution and a Gaussian. This is implemented in practice by drawing random samples from the distribution, that is, from the equilibrium MD simulation and adding the random noise vectors $N_i$. The neural network then attempts to reproduce the negative of the added noise, which is given by $r'-r$.

We can Taylor expand the density $\rho(r')$ about $r$ to the first order:
\[
\rho(r') = \rho(r) + (r'-r)\rho'(r),
\]
and substitute to get
\[
F_{\text{DN}} =\frac{\int_{-\infty}^{\infty} \left(r'-r\right) \left(\rho(r)+\left(r'-r\right)\rho'\left(r\right)\right)  \mathcal{N}(r,r',\sigma) dr'}{\int_{-\infty}^{\infty} \left(\rho(r)+(r'-r)\rho'(r)\right)  \mathcal{N}(r,r',\sigma) dr'}.
\]
Expanding out and cancelling the $\rho(r)$ from the top and bottom gives us
\[
F_{\text{DN}} =\frac{\int_{-\infty}^{\infty} (r'-r)\mathcal{N}(r,r',\sigma) dr' + \frac{\rho'(r)}{\rho(r)} \int_{-\infty}^{\infty} \left(r'-r\right)^2  \mathcal{N}(r,r',\sigma) dr'}{\int_{-\infty}^{\infty}  \mathcal{N}(r,r',\sigma) dr' +\frac{\rho'(r)}{\rho(r)} \int_{-\infty}^{\infty} \left(r'-r\right) \mathcal{N}(r,r',\sigma) dr'} 
\]
We have that
\[
\int_{-\infty}^\infty (r'-r)\mathcal{N}(r, r', \sigma)dr' = 0,
\]
as the integrand is an odd function. Hence,
\[
F_{\text{DN}} = \frac{\frac{\rho'(r)}{\rho(r)} \int_{-\infty}^{\infty} \left(r'-r\right)^2  \mathcal{N}(r,r',\sigma) dr'}{\int_{-\infty}^{\infty}  \mathcal{N}(r,r',\sigma) dr'} = \frac{\rho'(r)}{\rho(r)}\sigma^2.
\]
Now, the true energy at equilibrium is given by $E = -\log \rho(r)$, so the true forces are given by
\[
F_{\text{True}} = -\frac{dE}{dr} = \frac{d\log \rho(r)}{dr} = \frac{\rho'(r)}{\rho(r)}.
\]
We then arrive at
\[
F_{\text{DN}} = \sigma^2 F_{\text{True}}.
\]
That is, once we train a denoising model on the data pairs $(r_i+N_i, -N_i)$, scaling the outputs of the neural network by $1/\sigma^2$ gives a model that approximates the true forces on the system. 

The connection between the score and a force field, as well as its utility in generating molecular distributions, has been previously highlighted \citep{zaidi2022,arts2023}. However, a notable advantage of our derivation, apart from its simplicity, is that we can see what level of noise still reproduces the correct underlying distribution accurately. 
The noise level must be sufficiently small so that the perturbed samples remain within the region where the first-order Taylor expansion is still accurate.

If we instead expanded the Taylor series for $\rho(r')$ to the second degree, we arrive at the following approximation for $F_{\text{DN}}$:
\[
F_{\text{DN}}^{(2)} = \sigma^2F_{\text{True}}\left(1 + \frac{1}{2}\frac{\rho''(r)}{\rho(r)}\sigma^2\right)^{-1}.
\]
We can expand the denominator out as
\[
\left(1 + \frac{1}{2}\frac{\rho''(r)}{\rho(r)}\sigma^2\right)^{-1} = 1 - \frac{1}{2}\frac{\rho''(r)}{\rho(r)}\sigma^2 + \mathcal{O}(\sigma^4).
\]
Substituting this into the absolute error $\delta_{\text{DN}} = \left|F^{(2)}_{\text{DN}} + \sigma^2F_{\text{True}}\right|$, we get
\[
\delta_{\text{DN}} = \left| \frac{1}{2}\frac{\rho''(r)}{\rho(r)}\sigma^4 F_{\text{True}}\right| + \mathcal{O}(\sigma^6).
\]

\subsection{Time-dependent noise}
Note that in practice, there are several differences between MD and diffusion models. First, diffusion models employ a time-dependent score with larger noise levels at the initial stages of the process, so that the score converges to the true force field only toward the end of the generation process. This smoothing of the initial force field, analogous to thermal annealing, may accelerate convergence by preventing the model from becoming trapped in local energy minima. As a result, the actual trajectories will not exactly mirror those of conventional MD simulations until the noise level has been sufficiently reduced. 

This is also necessary as many diffusion models are trained only on minimum energy structures, which can be obtained from experiment (e.g., PDB structures). Therefore, they are in essence trained on zero temperature distributions. They hence cannot reproduce the real forces that would be experienced at higher temperatures. Thus, to generate training data far from the minimum energy structures it is essential to use the time-dependent noise.

\subsection{Coarse-grained MD} 
Additionally, most diffusion models cannot predict the full coordinates of every atom in a system, whereas most molecular dynamics simulations can, i.e., they employ an all-atom picture. However, it is also possible to perform coarse-grained MD simulations. Coarse-grained MD is a technique that simplifies complex molecular systems by reducing the degrees of freedom, enabling efficient simulation of large-scale behaviours while preserving essential physical properties. 

For example, only simulating the movement of a subset of the atoms, such as the atoms in a protein, while ignoring surrounding solvent molecules. This procedure is rigorously justified from a statistical mechanics perspective, and is known as coarse-grained MD. The distribution of ions is still given by the exponential of its free energy, i.e., a Boltzmann distribution. This correspondence allows us to treat the coarse-grained system with the standard tools of a neural network potential and molecular dynamics, even though certain degrees of freedom are ignored. For instance, a Langevin thermostat is frequently used, and the gradient of the free energy corresponds to the forces that result from averaging over the neglected degrees of freedom (i.e., the free energy is a potential of mean force). Next we will provide an example of this by running coarse-grained MD simulations of lithium chloride in water, with a diffusion model.

\section{Example: Lithium Chloride Solution}
One of the main benefits of using a diffusion model approach over a traditional NNP is that it only requires equilibrium coordinate data for training, rather than both coordinate and force data. While the mathematical demonstration that this is possible is significant, it remains unclear how much data is required for the diffusion model approach compared with training on forces. In this section, we use a simple example to illustrate the fundamental connection between denoising diffusion models and force recovery in MD. We first show how a network trained to predict the noise added to configurations extracted from MD recovers the underlying force field. Then, we investigate the effect of training data size on the stability of the model by comparing models trained on limited datasets against a traditional NNP trained on a larger set of force labels.

\begin{figure}[h]
    \centering
    \includegraphics[width=0.9\linewidth]{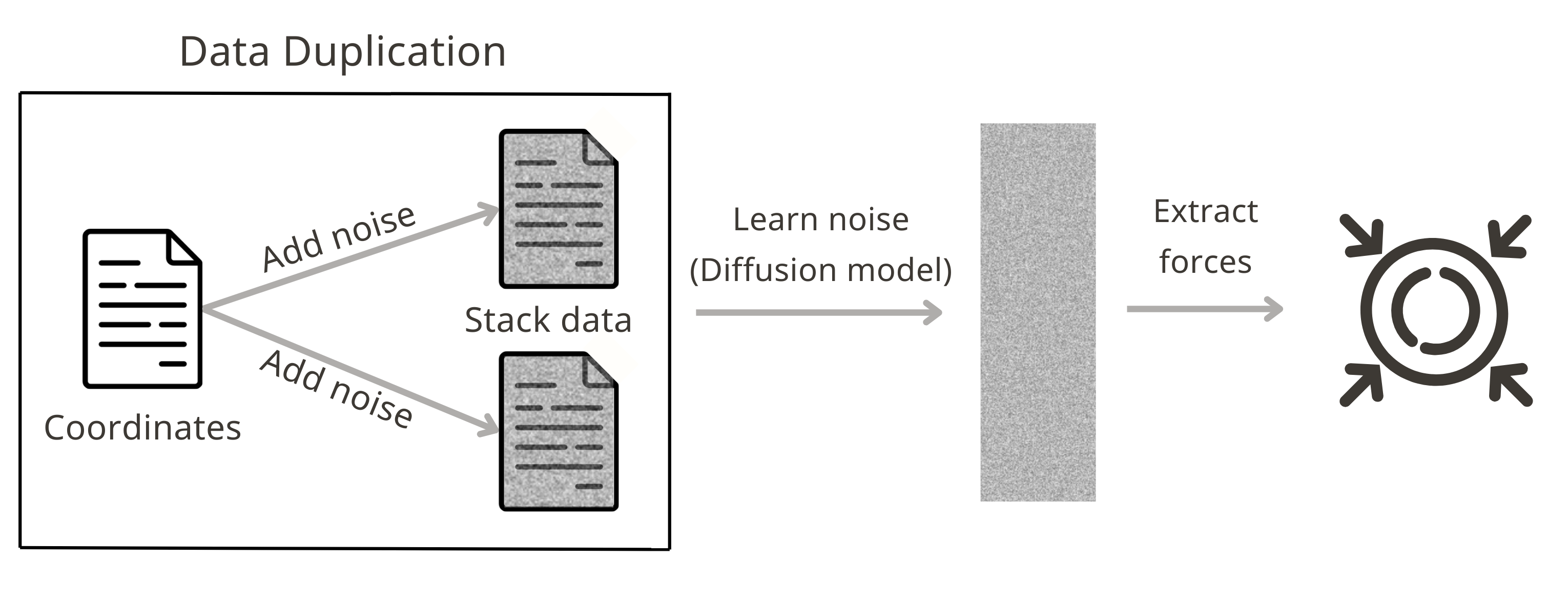}
    \caption{Pictorial description of the diffusion model and data duplication methodology.}
    \label{fig:data-duplication}
\end{figure}

The idea is to start with coordinate data frames $r_i$ (for $i=1, \dots, n$) taken from an all-atom MD simulation and add noise vectors $N_i$ taken from some distribution $\mathcal{N}$. We then train a denoising neural network to learn from the pairs $(r_i + N_i, -N_i)$. In fact, we can generate more noise vectors $M_i$ from $\mathcal{N}$ to create more training pairs $(r_i + M_i, -M_i)$ and stack these to create a training dataset twice as large. A flow diagram for this process is shown in Figure \ref{fig:data-duplication}.

\subsection{Methodology}
We chose to evaluate our model on its ability to simulate a coarse-grained lithium chloride solution in water at 300K, and we trained all our models using NequIP\citep{batzner2022}.

Here, we run coarse-grained MD of lithium chloride in water, meaning only the coordinates of the lithium and chloride ions are tracked, while the water molecules are omitted and their effects are represented through effective interactions. Consequently, the ions move on a free energy surface, where the solvent degrees of freedom have been integrated out, and the gradients of this free energy correspond to the average forces, i.e., the potential of the mean force.

We start with a traditional all-atom NNP simulating LiCl trained on high-level quantum chemical data, and perform MD simulations to generate coordinate and force data for a coarse-grained LiCl NNP. This process is detailed in the work by \citet{zhang2025}. We take 3000 frames of this coarse-grained coordinate and force data to train a benchmark NNP model to compare to our diffusion models, i.e., we train an NNP to predict only the average forces on the ions from the coordinates.  We then take 500 coordinate data frames $r_1, \dots, r_{500}$ and generate noise vectors $N_1, \dots, N_{500}$ and $M_1, \dots, M_{500}$ taken from a normal distribution with mean $r'$ and standard deviation $\sigma$. We train an augmented diffusion model using the $1000$ training pairs $(r_i + N_i, -N_i)$ and $(r_i + M_i, -M_i)$. Further, we take the 500 force vectors $F_i$ for each $r_i$ previously generated from the all-atom simulation and train a standard NNP using the 500 training pairs $(r_i, F_i)$.

Additionally, to demonstrate the utility of the data duplication technique, we train a diffusion model on the 500 training pairs $(r_i + N_i, -N_i)$. We will also check whether training the diffusion model on $1000$ unique coordinate frames is more beneficial than duplicating the same $500$ frames. To this end, we take an additional 500 coordinate frames $r_{501}, \dots, r_{1000}$, and train a diffusion model on the 1000 pairs $(r_i + N_i, -N_i)$.
All of our trained diffusion models have their outputs scaled by $1/\sigma^2$ as discussed in Section \ref{subsec:denoise}.

The idea is that 3000 frames represents the ideal training dataset size, while training a diffusion model on 500 frames simulates a scenario with limited available data. The results of our experiment will show that this efficiency, when combined with the diffusion training scheme, can provide a robust and effective framework for molecular dynamics in low-data settings.

We then use LAMMPS \citep{thompson2022} to run the molecular dynamics simulation for each model for a set amount of time, and compute the radial distribution functions (RDFs) produced from each run. The RDFs provide insight into the average spatial distribution of particles around a reference particle, revealing structural properties of the given system. This makes RDFs a useful tool for evaluating the qualitative stability of our models, as we can see if they are predicting the correct physical phenomena within the system.

\subsection{Results and Discussion}
\begin{figure}[ht]
    \centering
    \includegraphics[width=0.9\linewidth]{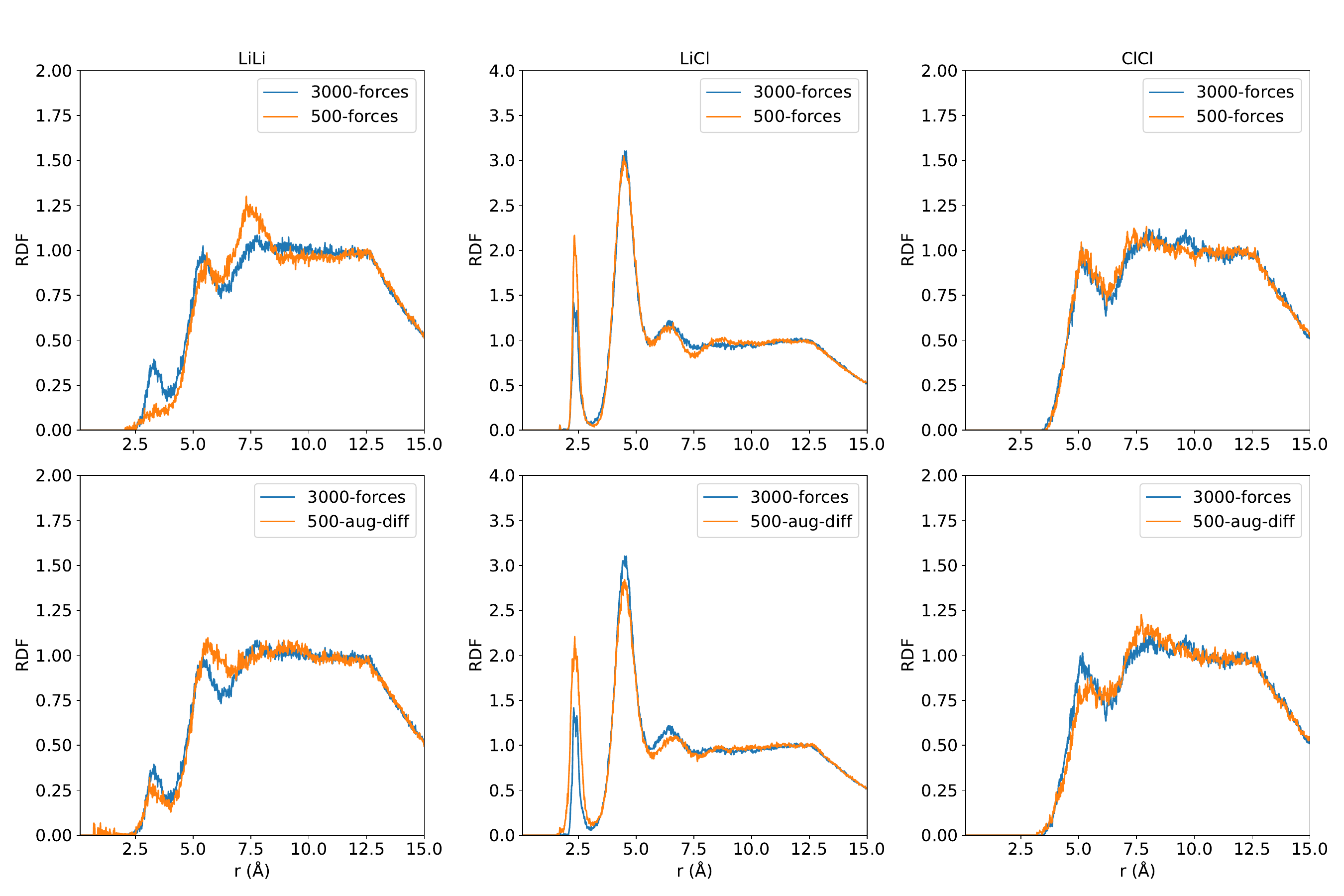}
    \caption{RDF plots for the NNP trained on 500 frames of force data (top row, 500-forces), and the augmented diffusion model (bottom row, 500-aug-diff), compared to the traditional NNP trained on 3000 frames of force data (3000-forces).}
    \label{fig:force-rdfs}
\end{figure}

We conduct the LAMMPS simulations for 50 ps (100,000 frames with a 0.5 fs timestep) for each model. The RDFs produced by the diffusion model trained on 500 frames of force data and the augmented diffusion model are shown in Figure \ref{fig:force-rdfs} against the RDFs produced by the benchmark NNP.

\begin{figure}[ht]
    \centering
    \includegraphics[width=0.9\linewidth]{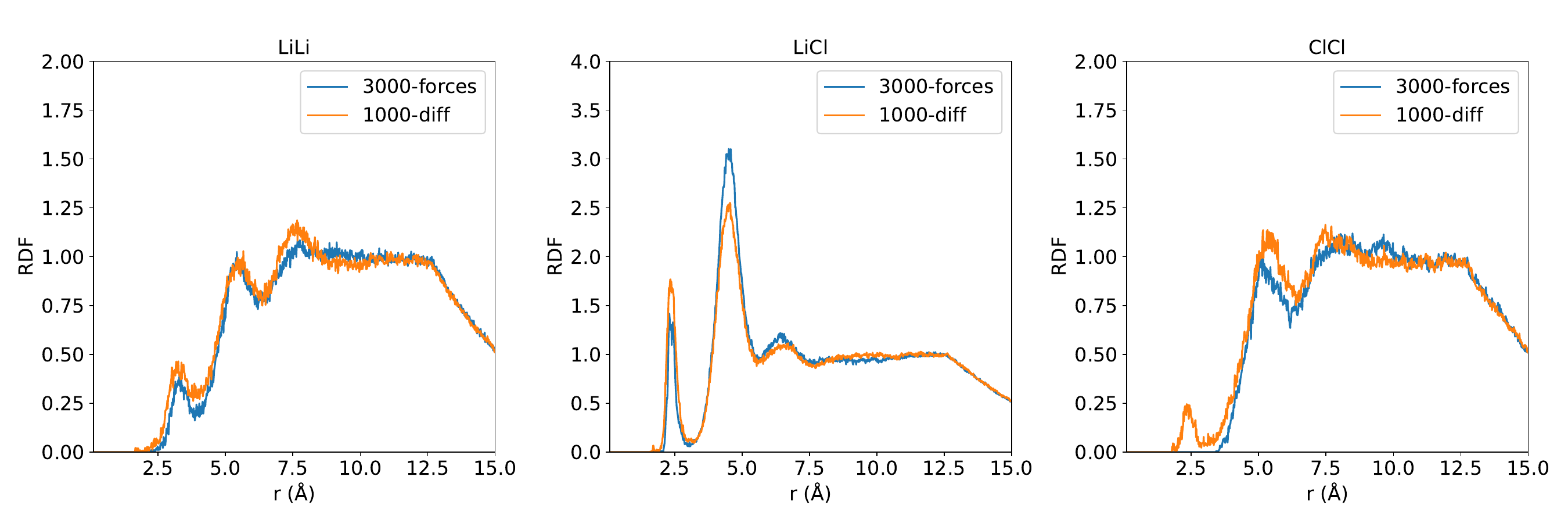}
    \caption{RDF plots for the regular diffusion model trained on 1000 frames of data compared to the benchmark traditional NNP.}
    \label{fig:1000-diff-rdfs}
\end{figure}

These plots demonstrate that the augmented diffusion  model produces results very similar to the 500-forces model, both of which match the RDFs produced by the benchmark 3000-forces model with slight inaccuracies. In particular, the augmented diffusion model accurately captures the structural properties of the system. This shows that by leveraging a diffusion-based training approach with an NNP and augmenting the dataset with additional noise vectors, we can match the performance of an NNP model trained on labelled force data. 

\begin{figure}[ht]
    \centering
    \includegraphics[width=0.9\linewidth]{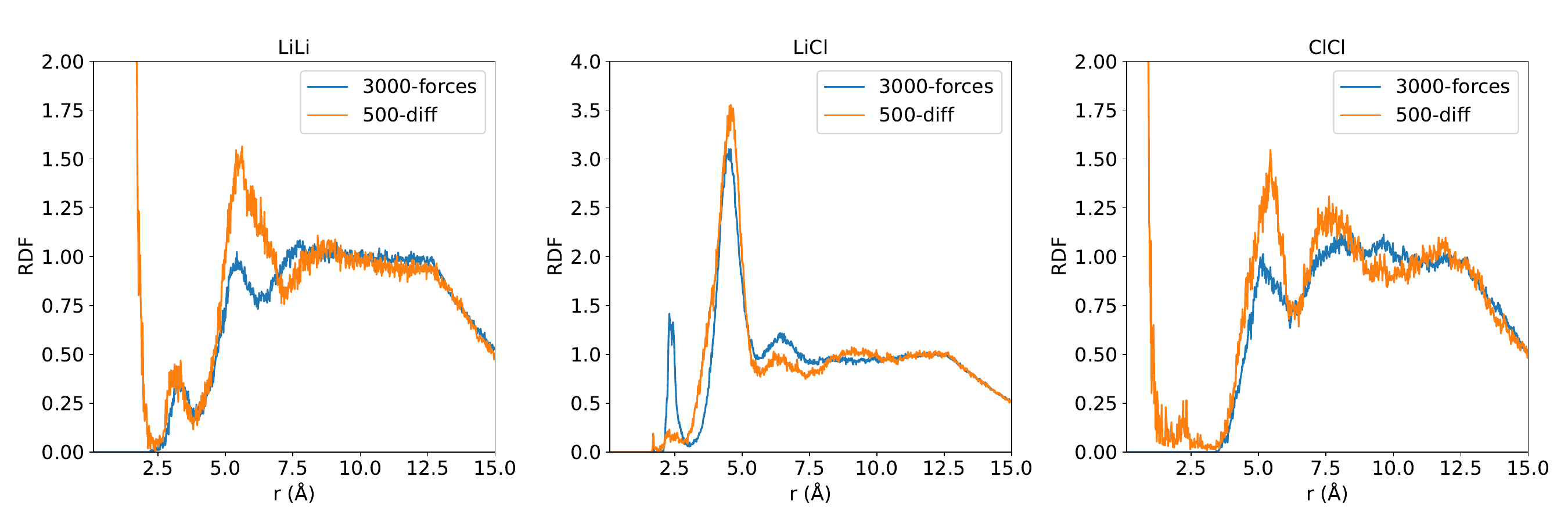}
    \caption{RDF plots for the regular diffusion model trained on 500 frames of data compared to the benchmark traditional NNP.}
    \label{fig:500-diff}
\end{figure}

Figure \ref{fig:1000-diff-rdfs} displays the RDFs produced by a diffusion model trained on 1000 frames of coordinate data (no duplication). A comparison with the augmented diffusion model (labeled 500-aug-diff in Figure \ref{fig:force-rdfs}) shows that both models yield similar accuracy with minor discrepancies observed. That is, there is no clear benefit in using 1000 unique coordinate data frames in the training process over duplicating 500 for the augmented model. However, the RDF plots for the diffusion model trained on only 500 frames of coordinate data without duplication (Figure \ref{fig:500-diff}) reveal that data duplication is essential when limited to 500 stable coordinate frames.

The large spikes in the Li-Li and Cl-Cl RDFs in the 0-2 \r{A} range show clear instability in the MD simulations of this model, which are absent in the benchmark NNP trained with force data. This indicates that the 500-diff model struggles to handle configurations in this small range, leading to unphysical and unstable behaviour in the simulations.

Our results confirm that the diffusion-based framework recovers forces in a manner consistent with traditional NNP approaches. Importantly, we demonstrate that with a given set of 500 coordinate data frames, we can duplicate the data and train a diffusion model to achieve comparable results to a traditional NNP trained on the same coordinate data with force labels.

\section{Conclusion}
In this work, we have provided a clear and straightforward demonstration of the intrinsic connection between diffusion models and molecular dynamics, both mathematically and with a practical example. 
By employing an NNP structure within a diffusion framework, we demonstrated that force fields can be recovered using only coordinate data, thereby eliminating the need for extensive force labels. This approach leverages standard MD and NNP software packages.
Moreover, we demonstrate that given a small amount of coordinate data, we can artificially expand a training dataset for a diffusion model twice as large whose stability in running MD simulations matches that of a traditional NNP trained on the same coordinate data but with force labels. These findings offer practical insights into the data requirements for achieving stable and accurate MD simulations in the low data settings.

\section{Future Work}
A potential avenue for future research is to systematically investigate how increasing the noise level affects the performance of the denoising model. By gradually raising the variance of the noise distribution, one could quantitatively evaluate whether the degradation of the model performance adheres to the error term derived from our Taylor expansion. This can provide valuable insights into the limits of the approximation and inform guidelines on selecting optimal noise levels.

Another direction entails a rigorous examination of training dataset configurations for the diffusion model. This would involve exploring various combinations of data duplication and possibly the integration of samples with true force labels in the dataset to determine the most effective strategy. Such an analysis, whether mathematical or experimental, would increase our understanding of how far this duplication technique can be stretched to perform stable molecular dynamics with minimal required training data.

\begin{acknowledgement}

This work was supported by resources provided by The Pawsey Supercomputing Centre with funding from the Australian Government and the Government of Western Australia. The authors also thank Kasimir Gregory for helpful discussions.

\end{acknowledgement}


\bibliography{ref}


\end{document}